\documentclass[letterpaper, 10pt, conference]{ieeeconf}

\IEEEoverridecommandlockouts                              
\usepackage{graphicx}
\usepackage{amsmath,amssymb,amsfonts}
\usepackage{xcolor}
\let\labelindent\relax
\usepackage{enumitem}
\usepackage{tabularx}
\usepackage{subcaption}
\usepackage{booktabs}
\usepackage{multirow,makecell}
\usepackage{bm}
\usepackage{hyperref}
\usepackage{csquotes}

\title{\LARGE \bf Modeling Output-Level Task Relatedness in Multi-Task Learning with Feedback Mechanism}
\author{Xiangming Xi, Feng Gao, Jun Xu, Fangtai Guo, and Tianlei Jin%
\thanks{Fangtai Guo is the corresponding author.}
\thanks{Xiangming Xi and Feng Gao are with the Intelligent Computing Infrastructure Innovation Center, Zhejiang Lab, Hangzhou, Zhejiang, China,  {\tt\small xixiangming@gmail.com, gaof@zhejianglab.com}. Fangtai Guo, and Tianlei Jin are with the Research Center for Intelligent Manufacturing Computing, Zhejiang Lab {\tt\small \{guofangtai, jtl\}@zhejianglab.com}. Jun Xu is with the School of Mechanical Engineering and Automation, Harbin Institute of Technology, Shenzhen, 518055, China,  {\tt\small xujunqgy@hit.edu.cn}.}
}
\begin{document}
%
\maketitle
\thispagestyle{empty}
\pagestyle{empty}
\begin{abstract}

Multi-task learning (MTL) is a paradigm that simultaneously learns multiple tasks by sharing information at different levels, enhancing the performance of each individual task. While previous research has primarily focused on feature-level or parameter-level task relatedness, and proposed various model architectures and learning algorithms to improve learning performance, we aim to explore output-level task relatedness. This approach introduces a posteriori information into the model, considering that different tasks may produce correlated outputs with mutual influences. We achieve this by incorporating a feedback mechanism into MTL models, where the output of one task serves as a hidden feature for another task, thereby transforming a static MTL model into a dynamic one. To ensure the training process converges, we introduce a convergence loss that measures the trend of a task's outputs during each iteration. Additionally, we propose a Gumbel gating mechanism to determine the optimal projection of feedback signals. We validate the effectiveness of our method and evaluate its performance through experiments conducted on several baseline models in spoken language understanding.
\end{abstract}
%

%
\section{Introduction}\label{sec-intro}

Multi-task learning (MTL) is a widely used deep learning framework that concurrently trains multiple related tasks by sharing information among them, aiming to enhance the performance of all tasks \cite{Zhang2021b}. Over the years, MTL has found successful applications in various fields, including spoken language understanding (SLU) \cite{Zhu2023b}, system identification \cite{Mahmoud2020}, prediction and control of landslide evolution \cite{Sun2022}, air traffic controller training \cite{Zhang2022}, etc. 

The effectiveness of MTL heavily relies on how information is extracted and utilized during training, with the design of the model architecture and the learning algorithm playing crucial roles. Existing literature often discusses shared information among tasks at either the feature level \cite{Argyriou2008} or the parameter level \cite{Zhang2018c}. Consequently, a task's output is primarily influenced by prior information, including task-specific and shared information represented as the hidden states of deep neural networks. However, in certain scenarios, task outputs may strongly depend on each other, indicating that posterior information could also be crucial for predicting other tasks.

Therefore, to delineate the relatedness of outputs among tasks and augment the capability of current MTL frameworks, we introduce a feedback mechanism that integrates one task's output into the learning process of another. This strategy imbues static MTL models with dynamism. Addressing two pivotal inquiries is imperative to stabilize the model and attain superior performance. Firstly, how can we guarantee convergent task predictions to enable the effective application of the learned model? Secondly, where should we integrate the task's output to facilitate the acquisition of more beneficial shared features?

In this paper, we aim to address these two inquiries. Our contributions are outlined as follows:
\begin{enumerate}[label=(\arabic*),itemindent=15pt,labelsep=3pt,fullwidth,parsep=0pt,labelwidth=0pt]
    \item We introduce a novel feedback mechanism into the MTL paradigm to describe task relatedness at the output level. This mechanism integrates one task's outputs as posterior information for the training/inference of another task.

    \item The incorporation of the feedback mechanism leads to a dynamic MTL model. We propose a loss function that assesses the convergence trend of task predictions in each iteration, ensuring the stability of the training process.

    \item To determine the optimal integration point for feedback information, we employ a gating mechanism based on the Gumbel distribution. This mechanism learns from data, identifying the most common features to extract and utilize.

    \item We conduct experiments on typical MTL models in spoken language understanding, comparing baseline models with and without the proposed feedback mechanism. Additionally, we present an ablation study on the convergence loss and Gumbel gating mechanism.
    
\end{enumerate}

    This paper is organized as follows. Section \ref{sec-review} briefly reviews the development of MTL concerning task relatedness and highlights the distinctions from our proposed method. Section \ref{sec-method} outlines the concept of the proposed feedback mechanism. Section \ref{sec-exp} details the comparative experiments conducted on various baseline models with/without the proposed feedback mechanism. Section \ref{sec-conclusion} concludes the paper.

\section{Literature Review}\label{sec-review}

\subsection{Learning Task Relatedness via MTL}

Multi-task learning has garnered attention from diverse communities for decades, primarily due to the widely accepted notion that related tasks can share valuable information during the learning process. To abstract and represent such shared information, researchers have proposed various approaches from different perspectives. As discussed in the survey by Zhang and Yang \cite{Zhang2021b}, existing approaches typically learn shared information either from the original or learned feature spaces or from the parameter spaces of different tasks.

Approaches focusing on the feature spaces often measure task relatedness through regularization of the original or learned features. For instance, Misra et al. \cite{Misra2016a} use the outputs of neurons in hidden layers for this purpose. On the other hand, approaches targeting the parameter spaces generally quantify task relatedness by analyzing the correlation or similarity of corresponding model parameters across different tasks. For example, Zhou and Zhao \cite{Zhou2016} cluster tasks by formulating and solving nonlinear optimization problems based on distance measures (such as the $\ell_2$ norm) of model parameters associated with different tasks.

Despite the differences in these approaches, it is worth noting that the learning and inference of MTL are generally "open-loop." This term, akin to the concept in control theory, signifies that channels for different tasks produce task-specific outputs conditioned solely on the model inputs, shared features, and task-specific features. However, task-specific outputs fail to provide posterior information during the training or inference of other tasks. This limitation can degrade model performance, particularly for tasks with strongly correlated output spaces. Thus, introducing output-level task relatedness to MTL is expected to enhance training and inference performance.

Moreover, some MTL models are designed based on practitioners' experience, incorporating unilateral correlation from one task to another \cite{Goo2018}, instead of learning symmetric correlations between tasks \cite{Zhao2021a}. While this approach may be effective in certain applications, it relies heavily on reliable prior information and may conflict with the ground truth.

In this paper, we aim to enhance the performance of MTL models by addressing the aforementioned challenges.

\subsection{Feedback Mechanism in Deep Learning}

The feedback mechanism is a fundamental concept in control theory, offering numerous benefits to various real-world applications. In the realm of deep learning, there are related techniques that leverage this concept. For instance, recurrent neural networks (RNNs) \cite{Wang2019h} and dynamic recursive neural networks (DRvNNs) \cite{Guo2019} are designed to incorporate the output of the model or component back into its input space.

RNNs consider both prior and posterior information, capturing local or contextual information of a token to generate its latent representation, typically for a single task. On the other hand, DRvNNs utilize backward connections to selectively neglect or reuse parts of the model, aiming to reduce model sizes and computational consumption or to enhance model adaptiveness, particularly in single-task learning scenarios. 

In contrast, our method focuses on capturing task relatedness by utilizing outputs as posterior information for the training and inference of other tasks.

\section{Methodology}\label{sec-method}
From a model architecture perspective, existing MTL models can be broadly categorized into two types, i.e., hard-parameter sharing and soft-parameter sharing \cite{Ruder2019}. These categories differ in the way they extract common features for different task channels.

In hard-parameter sharing, different tasks typically share a set of identical model parameters, such as encoder layers. On the other hand, in soft-parameter sharing, task models are physically independent, with common features learned using different regularization techniques on either feature representations or model parameters.

In both categorizations, tasks are primarily intertwined within layers from the input to hidden layers sequentially, with little consideration given to task outputs.

\begin{figure*}[htb]
        \begin{subfigure}[b]{0.45\textwidth}
             \centering
             \includegraphics[width=\columnwidth]{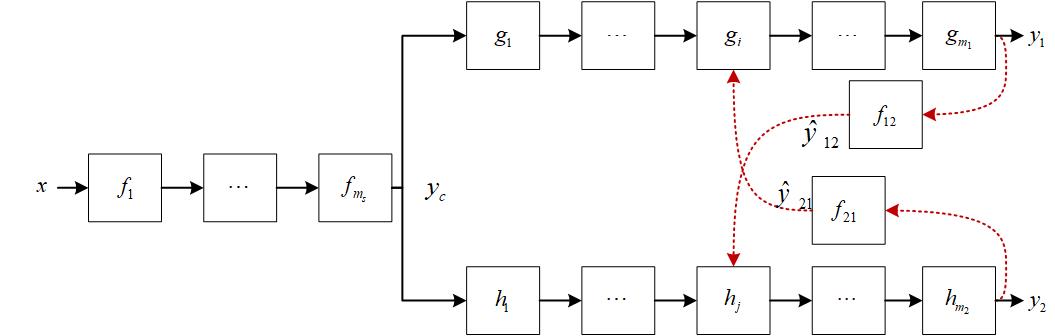}
             \caption{With fixed feedback infusion positions}
             \label{fig-framework-feedback}
         \end{subfigure}
        \begin{subfigure}[b]{0.45\textwidth}
             \centering
             \includegraphics[width=\columnwidth]{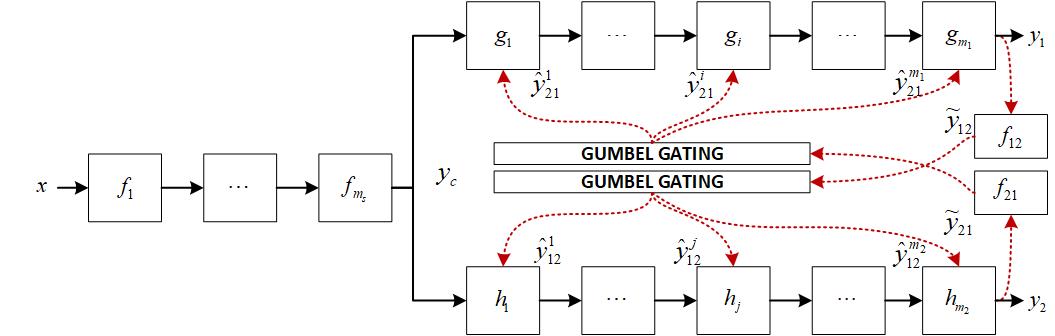}
             \caption{With the Gumbel gating mechanism}
             \label{fig-framework-gumbel}
         \end{subfigure}   
        \caption{Illustration of the proposed feedback and Gumbel gating mechanisms.}
        \label{fig-framework}
	\end{figure*}

    To model the output-level task relatedness, we propose a universal technique for MTL, i.e., the incorporation of the feedback mechanism. Without loss of generality, we take a hard-parameter sharing MTL model with two tasks as an example and illustrate the mechanism in Fig. \ref{fig-framework}. The blocks, $\{ f_i\}_{i=1}^{m_s}$, $m_s \in \mathbb{Z}_{+}$, denote the shared layers of two tasks, i.e., Task 1 and Task 2. When $m_s = 0$, it implies that the two tasks share no layers. Blocks $\{g_i\}_{i = 1}^{m_1}$ and $\{h_i\}_{i = 1}^{m_2}$, $m_1, m_2 \in \mathbb{Z}_{++}$, denote the task-specific layers. Given input $\bm{x} \in \mathbb{R}^n$, $\bm{y}_1 \in \mathcal{Y}_1$ and $\bm{y}_2 \in \mathcal{Y}_2$ are the outputs of the two related tasks.

    To model the output of Task 1, denoted by $\bm{y}_1$, as posterior information for Task 2, we branch and infuse it via an amplifier block, denoted by $f_{12}$, as an extra input into the $j$-th block in the task-specific layers of Task 2, i.e., $h_j$, $1 \leq j \leq m_2$, as is shown with the red dashed line in Fig. \ref{fig-framework-feedback}. Similar operations are also applied to the output of Task2, denoted by $\bm{y}_2$, and block $g_i$, $1 \leq i \leq m_1$ with an amplifier block $f_{21}$. These amplifier blocks are necessary since the representation of task outputs might not align with each other. The subscript in $f_{ij}$ indicates that the feedback information is from Task $i$ to Task $j$. Therefore, the outputs of the model in Fig. \ref{fig-framework-feedback} can be calculated as follows,
    	\begin{equation}
        \left\{
			\begin{array}{l}
				\bm{y}_1^{k} = g_{m_1} \circ \cdots \circ g_{i} (\hat{\bm{y}}_{21}^{k-1}, g_{i-1} \circ \cdots \circ g_{1} (\bm{y}_c)),\\
				\bm{y}_2^{k} = h_{m_1} \circ \cdots \circ h_{j} (\hat{\bm{y}}_{12}^{k-1}, h_{j-1} \circ \cdots \circ h_{1} (\bm{y}_c)),\\
				\bm{y}_c = f_{m_s} \circ \cdots \circ f_1  (\bm{x}),\\
                \hat{\bm{y}}_{21}^{k-1} = f_{21}(\bm{y}_2^{k-1}),\\
                \hat{\bm{y}}_{12}^{k-1} = f_{12}(\bm{y}_1^{k-1}),
			\end{array}\right.
		\end{equation}
    where $\bm{y}_c$ is the output of the shared layers, $\circ$ represents function composition, and the superscript $k \in (0, K]$, $K \in \mathbb{Z}_{++}$, indicates the iteration step during a single prediction, $\bm{y}_{1}^{0},\bm{y}_{2}^{0}$ are initialized task outputs. We adopt the last prediction, i.e., $\bm{y}_{1}^{K}$ and $\bm{y}_{2}^{K}$, as the final task predictions.

    Associated with the feedback mechanism, three core components need to be considered for better performance.

    \begin{enumerate}[label=(\arabic*),itemindent=15pt,labelsep=3pt,fullwidth,parsep=0pt,labelwidth=0pt]

        \item The first is how to determine where task outputs should be infused. An intuitive method is to take it as a hyperparameter determined with cross-validation or other techniques. An alternative is to apply the gating technique. Motivated by the gating technique utilized in \cite{Guo2019} to reduce model sizes, we propose to utilize the continuous relaxation of the Gumbel-Max trick based on Gumbel distribution as a control of the infusion of feedback signals. As is shown in Fig. \ref{fig-framework-gumbel}, given the feedback amplifier $f_{ij}$ from Task $i$ to Task $j$, the gate threshold vector is computed as follows,   
        \begin{equation}
            \bm{\gamma}_{ij} = \left\{
            \begin{array}{ll}
                \bm{e}_{\hat{t}}, & \mbox{in forward pass},  \\
                \mbox{softmax}(\Tilde{y}_{ij}^{\rm MAX} + \bm{G}), & \mbox{in backward pass},
            \end{array}\right.
        \end{equation}
        where $\bm{e}_{\hat{t}} \in \mathbb{R}^{m_j}$ is the unit vector with the $\hat{t}$-th entry being 1, $\hat{t} = \mbox{argmax}_{t}(\Tilde{y}_{ij}^{\rm MAX} + G_t)$, $\Tilde{y}_{ij}^{\rm MAX} = ||\Tilde{\bm y}_{ij}||_{\infty}$, $\bm{G} = (G_1, \cdots, G_{m_j})^T$ is a Gumbel random vector of i.i.d., and $\Tilde{\bm y}_{ij} = f_{ij} (\bm y_{i})$ is the amplified task output of Task $i$ for Task $j$. Therefore, the feedback signal for the $t$-th block in the task-specific layer of Task $j$ in the $k$-th iteration is as follows,
        \begin{equation}
            \hat{\bm y}_{ij}^{k, t} = \bm{\gamma}_{ij}^{(k, t)} \cdot \Tilde{\bm y}_{ij}.
        \end{equation}

        \item The second is the measure of stability of task predictions, i.e., whether the sequence of predicted outputs of a specific task is convergent. Denote the sequential prediction of Task $i$ by $\bm{y}_i^{0}, \cdots, \bm{y}_i^{K}$. As dynamic systems are expected to be convergent to a steady state, we are supposed to obtain
        \begin{equation}
            \lim_{k \to \infty} ||\bm{y}_i^{k+1} - \bm{y}_i^{k}||_\mathcal{F} = 0.
        \end{equation}
        Therefore, to measure the convergence of the output sequence, we propose a novel convergence loss function, i.e., 
        \begin{equation}\label{eqn-loss}
            L(\{\bm{y}_i^{k}\}_{k=0}^K) = \sum_{k = 1}^{K-1} \beta^{K-k} ||\bm{y}_i^{k+1} - \bm{y}_i^{k}||_\mathcal{F}
        \end{equation}
        where $\beta \in (0, 1)$ is a delay constant that controls the forget degree of previous predictions. In general, the loss function (\ref{eqn-loss}) can be combined with other commonly used loss functions during the learning of model parameters.

        \item The third is the determination of the hyperparameter $K$ in (\ref{eqn-loss}). Since the MTL model with the feedback mechanism is dynamic, it is expected to converge to a steady state along the time dimension. However, in most real-world applications, a prediction is supposed to be made within a very short period. This results in the trade-off between convergence and efficiency for predictions.
    \end{enumerate}

It is straightforward to generalize the proposed method for three or more tasks by incorporating the output of a single task into the pass of another. In the extreme case where any two tasks are related concerning their outputs, the connection complexity might increase exponentially as the number of tasks increases. However, this might be uncommon, and other techniques such as model pruning \cite{Menghani2023} might be utilized to remove unnecessary connections.

\section{Experiments}\label{sec-exp}

In this section, we incorporate the proposed method into several MTL models in SLU and evaluate their performances with a brief analysis. 

The tasks used in this section include intent detection (ID), slot filling (SF), and next-word prediction (NWP). Here is a simple example \cite{Liu2016h}. Given a user utterance, i.e.,

\begin{displayquote}
    \textit{Show flights from Seattle to San Diego tomorrow}
\end{displayquote}

\begin{itemize}
    \item Task \textit{ID} recognizes the user's intent by expressing the above utterance, e.g., \textit{Flight}. It is often modeled as a classification problem.

    \item Task \textit{SF} predicts what kind of information each token in the utterance implies, and is modeled as a sequence labeling problem. For the above example, the ground-truth labels under the IBO annotation framework are ``\textit{O/O/O/B-fromloc/O/B-toloc/I-toloc/B-depart\_date}''.

    \item Task \textit{NWP} predicts the next token based on previous tokens, e.g., to maximize $P(from \mid show,~flights)$.
\end{itemize}

\subsection{Settings}

\begin{figure}[tb]
        \begin{subfigure}[b]{0.24\textwidth}
             \centering
             \includegraphics[width=\columnwidth]{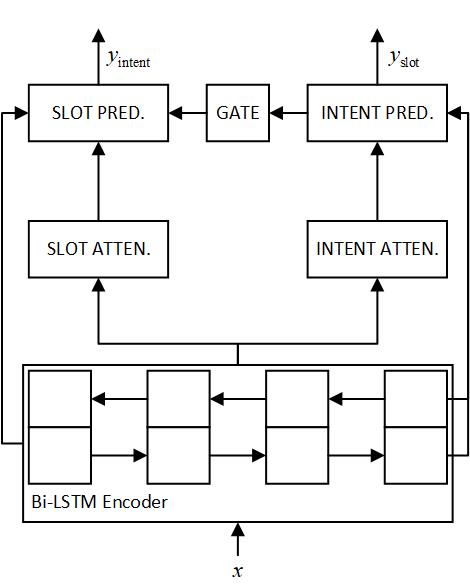}
             \caption{}
             \label{fig-ALG_DEMO-gate-ori}
         \end{subfigure}
        \begin{subfigure}[b]{0.24\textwidth}
             \centering
             \includegraphics[width=\columnwidth]{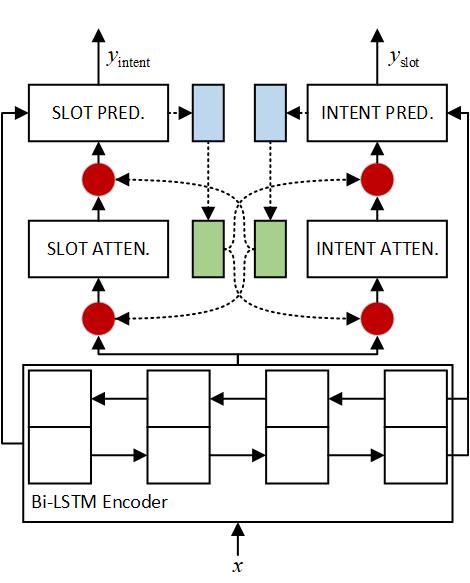}
             \caption{}
             \label{fig-ALG_DEMO-gate-feedback}
         \end{subfigure}
        \caption{Illustration of the SG model (a) and its counterpart with the proposed feedback mechanism (b). Blue blocks represent feedback amplifiers, green blocks represent Gumbel gating blocks, red circles indicate candidate infusion positions, and dashed lines indicate potential feedback routes.}\label{fig-ALG_DEMO}
    \end{figure}

\begin{figure}[tb]
        \begin{subfigure}[b]{0.24\textwidth}
             \centering
             \includegraphics[width=.8\columnwidth]{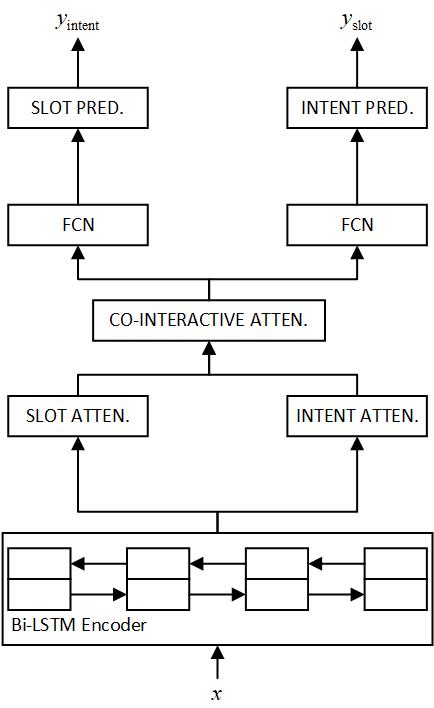}
             \caption{}
             \label{fig-ALG_DEMO-dca-ori}
         \end{subfigure}
        \begin{subfigure}[b]{0.24\textwidth}
             \centering
             \includegraphics[width=\columnwidth]{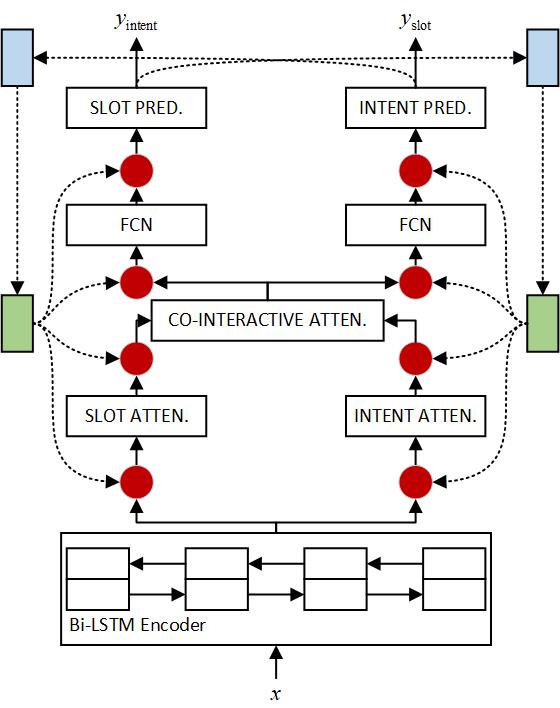}
             \caption{}
             \label{fig-ALG_DEMO-dca-feedback}
         \end{subfigure}
        \caption{Illustration of the CIT model (a) and its counterpart with the proposed feedback mechanism (b). }\label{fig-ALG_DEMO-cit}
    \end{figure}

\begin{figure}[tb]
        \begin{subfigure}[b]{0.24\textwidth}
             \centering
             \includegraphics[width=\columnwidth]{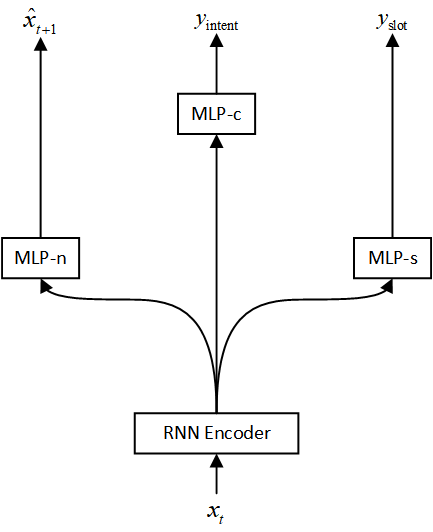}
             \caption{}
             \label{fig-ALG_DEMO_liu-ori}
         \end{subfigure}
        \begin{subfigure}[b]{0.24\textwidth}
             \centering
             \includegraphics[width=\columnwidth]{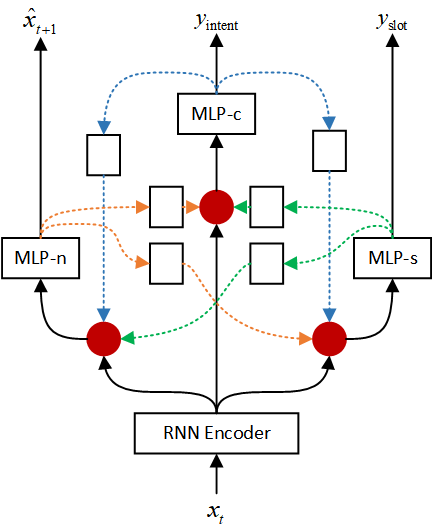}
             \caption{}
             \label{fig-ALG_DEMO_liu-feedback}
         \end{subfigure}
        \caption{Illustration of the SLU-LM mode \cite{Liu2016h} (a) and its counterpart with the proposed feedback mechanism (b). The blue, green, and orange dashed lines are the feedback routes originating from Task ID, SF, and NWP, respectively.}\label{fig-ALG_DEMO_liu}
    \end{figure}

    Considering that the proposed feedback mechanism is universal, we select three typical MTL models concerning different numbers of tasks and incorporate the proposed feedback mechanism both into the architecture and into the training process. The selected models include,
    \begin{enumerate}[label=(\arabic*),itemindent=15pt,labelsep=3pt,fullwidth,parsep=0pt,labelwidth=0pt]
        \item the slot-gated (SG) model \cite{Goo2018}, which jointly considers ID and SF under the hard-parameter sharing framework. The slot-gate mechanism models the unilateral interaction from ID to SF and is shown in Fig. \ref{fig-ALG_DEMO-gate-ori};

        \item the Co-Interactive Transformer (CIT) \cite{Qin2021}, which models the mutual influence between ID and SF, as shown in Fig. \ref{fig-ALG_DEMO-dca-ori}. 

        \item the joint model of spoken language understanding and language model (SLU-LM) \cite{Liu2016h}, which considers ID, SF, and NWP simultaneously, as shown in Fig. \ref{fig-ALG_DEMO_liu-ori}; 

    \end{enumerate}
    
    For the baseline models, we utilize the open-sourced implementation of OpenSLU\footnote{\url{https://github.com/LightChen233/OpenSLU}} \cite{Qin2023} (for SG and CIT)  and SLU-LM\footnote{\url{https://github.com/HadoopIt/joint-slu-lm}}. By modifying their architectures to the minimum, we integrate the feedback mechanism into each baseline model, and the resulting model architectures are illustrated in Fig. \ref{fig-ALG_DEMO-gate-feedback}, \ref{fig-ALG_DEMO-dca-feedback}, and \ref{fig-ALG_DEMO_liu-feedback}, respectively. We compare all models with their default configurations, including batch sizes, accelerator types, learning rates, etc.

    We employ the widely-used ATIS dataset \cite{Dahl1994} following the experimental setup established in \cite{Liu2016h}. The dataset statistics are summarized in Table \ref{tbl::dataset}. We employ metrics consistent with prior studies to facilitate a comprehensive comparison across multiple tasks, i.e., ID, SF, and NWP. Specifically, we record all models' intent accuracy (ACC) and slot F1-score. For NWP, we assess the language model perplexity (PPL) to gauge a model's predictive performance for the next token, computed as the exponentiation of the cross-entropy between the data and model predictions.
    Additionally, for SG and CIT, we evaluate overall task performance concerning ID and SF using the exact match accuracy (EMA), which quantifies the accuracy of correctly predicting both intent and all slot labels. To evaluate the proposed convergence loss, we utilize the setting steps, defined as the training step where a specific metric reaches and maintains within a certain range of its final value. In our case, we set the range to 2\%.

\begin{table}[]
\centering
\caption{Statistics of the ATIS dataset.}\label{tbl::dataset}
\small
\begin{tabular}{@{}lllll@{}}
\toprule
      & \#INTENT & \#SLOT & \#TRAIN & \#TEST \\ \midrule
ATIS  & 12       & 127    & 4978    & 893    \\\bottomrule
\end{tabular}
\end{table}

\subsection{Effectiveness}
In this section, we compare the three baseline models with their counterparts incorporating the proposed feedback mechanism to validate its effectiveness in enhancing task relatedness and improving performance. The feedback mechanism is configured with $K = 4$ and randomly initialized task outputs. We use the weighted sum of the cross-entropy and convergence loss as the training objective function with adaptive weights as in \cite{He2021a}.

We present the experiment results in Table \ref{tbl-exp1-openslu} and \ref{tbl-exp1-slu-lm}, where the metrics for the baseline models are obtained from the corresponding references.

\begin{table}[t]
\caption{Performance of SG and CIT with/without the feedback mechanism.}
\label{tbl-exp1-openslu}
\resizebox{\columnwidth}{!}{%
\begin{tabular}{@{}cl|lll@{}}
\toprule
\# & Method          & \begin{tabular}[c]{@{}l@{}} Intent/ACC \\(\%)\end{tabular}  & \begin{tabular}[c]{@{}l@{}} Slot/F1 \\(\%)\end{tabular}        & EMA (\%) \\\midrule
1 & SG w/o feedback  & 94.5            & 94.7               & 82.5     \\
2 & SG w/ feedback   & 96.8            & 94.9               & 84.2     \\\hline
3 & CIT w/o feedback & 97.3            & 95.9               & 87.6     \\
4 & CIT w/ feedback  & 97.9            & 96.3               & 88.1     \\\bottomrule
\end{tabular}%
}
\end{table}

\begin{table}[t]
\caption{Performance of SLU-LM with/without the feedback mechanism. }
\label{tbl-exp1-slu-lm}
\small
\begin{tabular}{@{}cl|lll@{}}
\toprule
\# & Method                 & \begin{tabular}[c]{@{}l@{}} Intent/ACC \\(\%)\end{tabular}  & \begin{tabular}[c]{@{}l@{}} Slot/F1 \\(\%)\end{tabular}   & LM/PPL \\ \midrule
1 & BASIC                   & 97.98      & 94.15         & 11.33  \\\hline
2 & LI     w/o feedback     & 98.10      & 94.22         & 11.27  \\
3 & LI     w/ feedback      & 98.30      & 94.29         & 11.05  \\\hline
4 & LS     w/o feedback     & 98.21      & 94.14         & 11.14  \\
5 & LS w/ feedback          & 98.32      & 95.30         & 10.50  \\\hline
6 & LS \& LI w/o feedback   & 98.10      & 94.13         & 11.22  \\
7 & LS \& LI w/ feedback    & 98.42      & 98.80         & 9.33  \\ \bottomrule
\multicolumn{5}{p{230pt}}{Note: \textit{BASIC} is the SLU-LM model without any feedback route, as shown in Fig. \ref{fig-ALG_DEMO_liu-ori}. \textit{LI}/\textit{LS} indicates that the feedback route from ID/SF to NWP is presented in the model. The \textit{feedback} version of each method implies that the vice versus feedback route is included, i.e., from NWP to ID/SF.} 
\end{tabular}%
\end{table}

From the results in Table \ref{tbl-exp1-openslu}, it is evident that the incorporation of the proposed feedback mechanism enhances the performance of both tasks compared to the baseline models (SG and CIT), which consist of two correlated tasks. Notably, there is a more significant improvement in Intent/ACC compared to Slot/F1. This observation can be attributed to the nature of the tasks: Task ID is a sentence-level task that relies heavily on global context information, whereas Task SF is a token-level task that primarily utilizes local context information over high-level features.

As to SLU-LM, which has three sub-tasks, we have the following observations in the results in Table \ref{tbl-exp1-slu-lm}. First, Methods 3 and 5 improve the performance of Intent/ACC and Slot/F1 more significantly, respectively. Though there is a feedback route in Method 2/4 from Task ID/SF to NWP, it is still a static model that takes the features of ID/SF as a contributing component to Task NWP. When introducing the additional feedback route from Task NWP back to ID/SF, as in Method 3/5, the dynamic nature of the model has a positive effect on both tasks. Second, from the results for Methods 2/3 and 4/5, the improvements are more significant in the latter pair. It implies that Task SF and NWP are more correlated since they are both token-level tasks. Third, for Method 6/7, the performance of all three tasks improves and is better than that in Method 2-5, which proves the correlation among the three tasks.

\subsection{Alation Study}
\begin{figure}[h]
        \begin{subfigure}[b]{0.5\textwidth}
             \centering
             \includegraphics[width=\columnwidth]{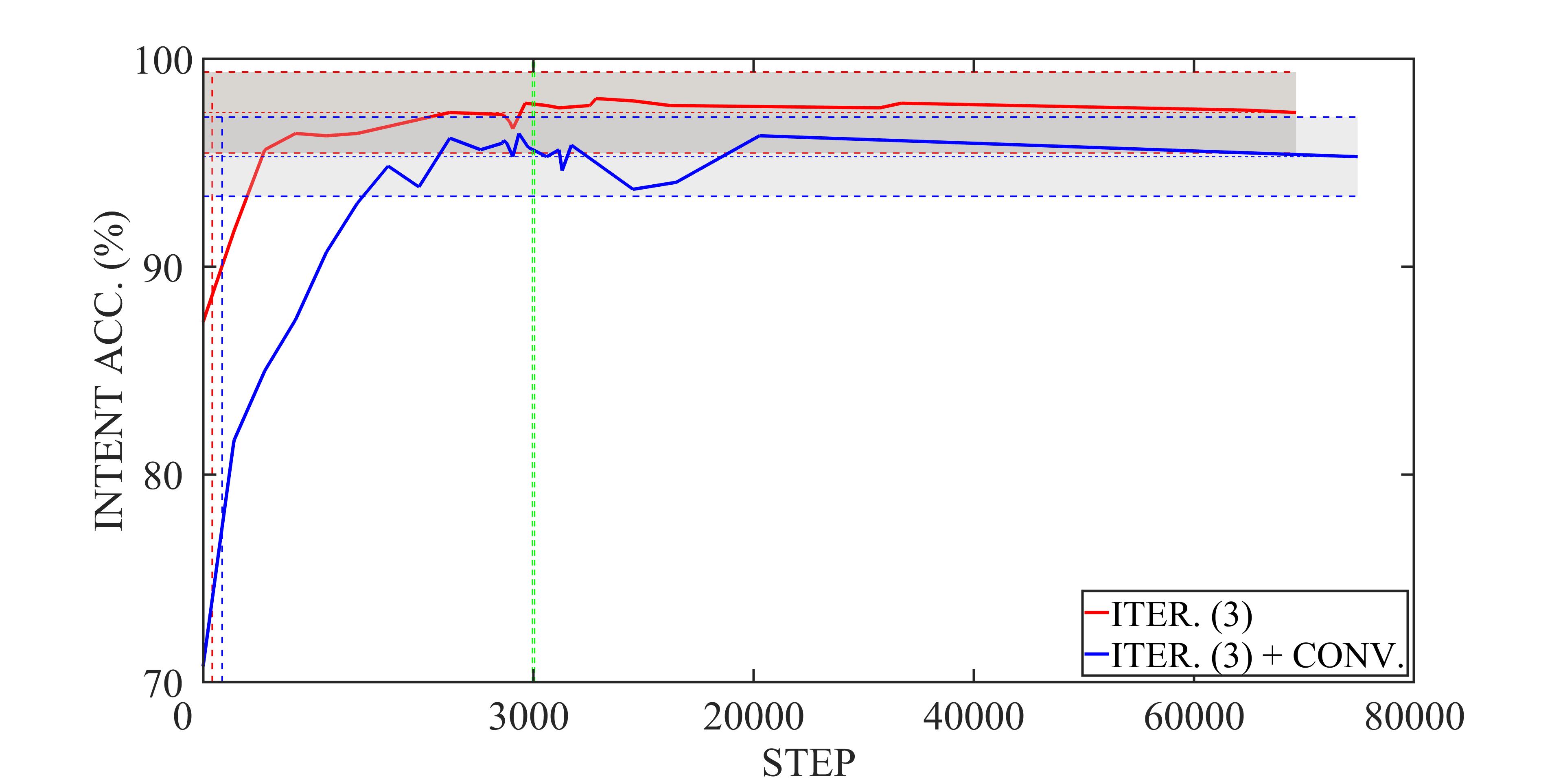}
             \caption{Intent accuracy}
             \label{fig-res-intent}
         \end{subfigure}
        \begin{subfigure}[b]{0.5\textwidth}
             \centering
             \includegraphics[width=\columnwidth]{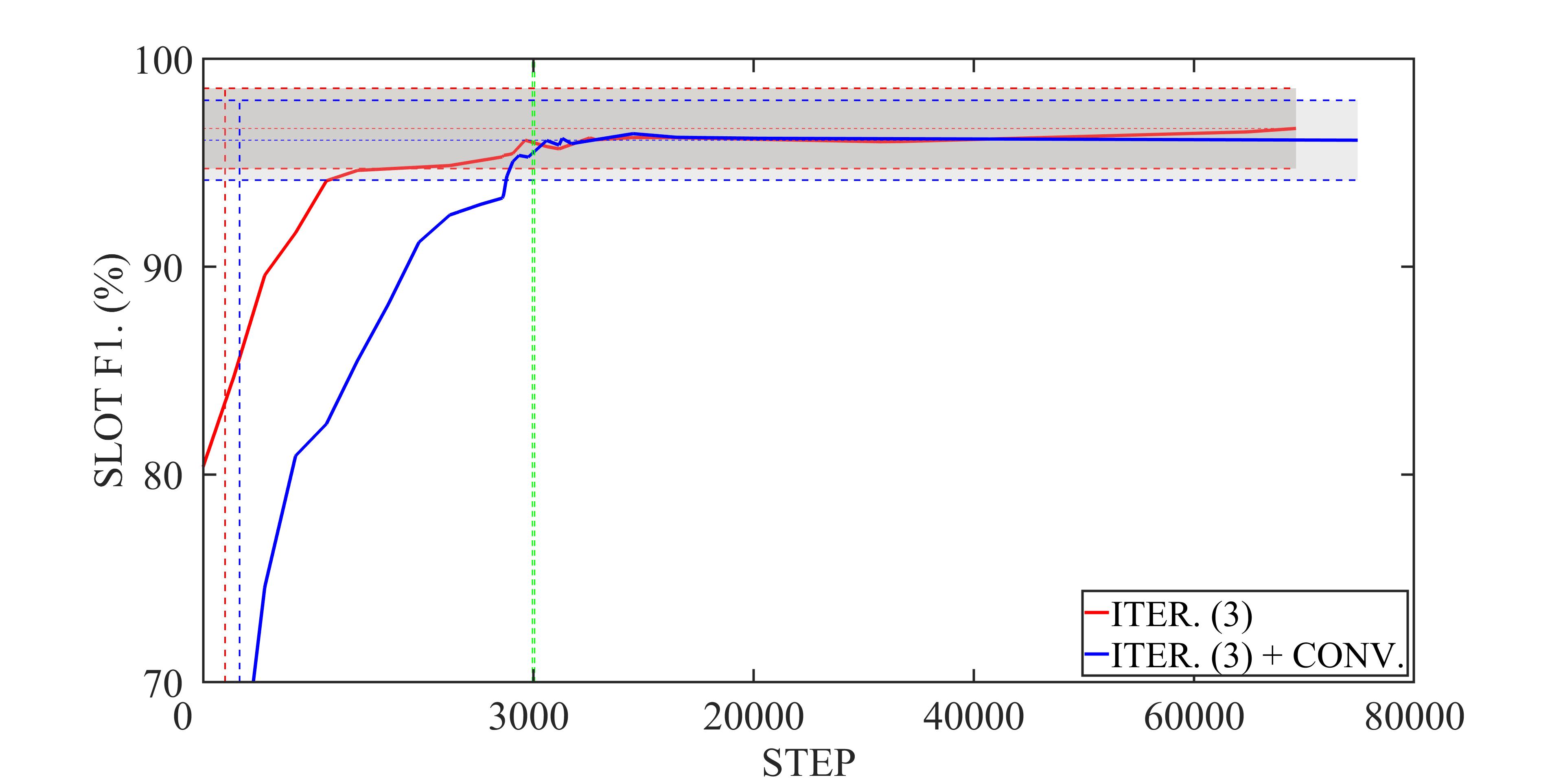}
             \caption{Slot F1}
             \label{fig-res-slot}
         \end{subfigure} 
\caption{Performance comparison of CIT (ITER.) and CIT (ITER. + CONV.). For better visualization, we scale the steps between 0 and 3000 10x, With setting steps 828/1732 for \textit{Intent/ACC}  and 1995/3311 for \textit{Slot/F1}, respectively.}
\label{fig-res}
\end{figure}

In this section, we delve into the importance of the convergence loss and the Gumbel gating mechanism through an ablation study on the SG and CIT models. Additionally, we configure the feedback mechanism with randomly initialized task outputs. The results of these experiments are summarized in Table \ref{tbl-ablation}. Furthermore, we explore the impact of the feedback mechanism on the convergence of the training process, with the findings illustrated in Fig. \ref{fig-res}.


   From Table \ref{tbl-ablation},  we find that the performance of both tasks in SG (TRIV.) degrades compared to that of Method 1 in Table \ref{tbl-exp1-openslu}. This is because SG (TRIV.) breaks the unilateral interaction from ID to SF for the incorporation of the feedback mechanism, as shown in Fig. \ref{fig-ALG_DEMO}. And it confirms the positive improvements of sharing common features in the MTL paradigm. On the contrary, the difference between CIT (ITER.) and Method 3 in Table \ref{tbl-exp1-openslu} is not significant, since the modification to CIT does not alter the model architecture, as shown in Fig. \ref{fig-ALG_DEMO-cit}. 

   Comparing SG/CIT (ITER.) with SG/CIT (TRIV.), the performance of both tasks, as well as the overall performance, EMA, all improves, which shows that the utilization of the feedback information alone can result in benefits to the model performance. Though the utilization of the convergence loss can bring improvements in Intent/ACC as shown in SG/CIT (ITER. + CONV.), the performance on Slot/F1 and EMA degrades, compared to SG/CIT (ITER.). An explanation is that the convergence loss emphasizes the local steady states of tasks during one iteration, and has a side effect on better search in the parameter space towards the optimality. Besides, an improvement in all metrics can be observed in SG/CIT (ITER. + GG.), compared to SG/CIT (ITER.). This implies that instead of fixing the position where the feedback information is infused, learning such behavior based on the Gumbel gating mechanism is of more significance. Further, when integrating all the techniques, i.e., SG/CIT (Ful.), it outperforms all the other methods, and it validates the effectiveness of the proposed mechanism.

    We visualize the feedback loss values of Task ID/SF for CIT (ITER.) and CIT (ITER. + CONV.) in Fig. \ref{fig-res}. The figure indicates that methods incorporating the convergence loss (\textit{CONV.}) exhibit slower convergence compared to those without it. This behavior can be attributed to the fact that the proposed convergence loss places greater emphasis on reducing the variance of sequential predictions within one iteration, thereby leading to a slower overall convergence rate. However, a notable benefit of this approach is that both tasks tend to be more robust against noise during testing.

\begin{table}[t]
\caption{Ablation study on the baseline models with the feedback mechanism under different configurations.}\label{tbl-ablation}
\centering
\small
\resizebox{\columnwidth}{!}{%
\begin{tabular}{@{}cllll@{}}
\toprule
\# & Method & \begin{tabular}[c]{@{}l@{}} Slot/F1 \\(\%)\end{tabular}   &  \begin{tabular}[c]{@{}l@{}} Intent/ACC \\(\%)\end{tabular}  & \begin{tabular}[c]{@{}l@{}} EMA \\(\%)\end{tabular}\\ \midrule %
1 & SG (TRIV.)            &  83.9  &   94.4  & 80.0 \\
2 & SG (ITER.)            &  86.1  &  95.4   & 81.5 \\
3 & SG (ITER. + CONV.)    &  84.1  & 96.4    & 81.3 \\
4 & SG (ITER. +  GG.)     &   90.2 &  95.8   & 82.0 \\
5 & SG (FUL.)             &  94.9  & 96.8    & 84.2  \\\midrule 
6 & CIT (TRIV.)           &  96.0  &  97.3   & 87.5 \\
7 & CIT (ITER.)           &   96.2 &  97.4   &  87.6 \\
8 & CIT (ITER. + CONV.)   &  96.1  &  97.9   & 87.6 \\
9 & CIT (ITER. +  GG.)    &   96.3 & 97.5    & 86.7  \\
10 & CIT (FUL.)            & 96.3   &   97.9  & 88.1 \\\bottomrule 
\multicolumn{5}{p{230pt}}{Note: \textit{TRIV.} indicates the baseline model has been modified to incorporate the feedback mechanism, but no feedback information is used with $K = 1$. \textit{ITER.} indicates the utilization of the feedback mechanism, i.e., $K > 1$. \textit{CONV.} indicates the utilization of the proposed convergence loss. \textit{GG.} indicates that the Gumbel gating mechanism is used, while for methods without GG., the feedback point is fixed and determined by cross-validation. \textit{FUL.} is short for ITER. + CONV. + GG.}   \\ 
\end{tabular}
}
\end{table}

\section{Conclusions}\label{sec-conclusion}
Mlti-task learning (MTL) has emerged as a powerful paradigm for simultaneously learning multiple related tasks by extracting and sharing common features. However, existing MTL approaches often model task relatedness at the feature or parameter level, emphasizing more on prior information. In this paper, we explore the output-level task relatedness. Specifically, we introduce the well-accepted feedback mechanism into MTL, which leverages the output of one task as posterior information for other tasks. 

Since the feedback mechanism imbues dynamism into a static MTL model, we need to address the problem of stabilizing the prediction of each task in one iteration and determining the optimal infusion of feedback signals. To ensure the convergence of each task, we utilize a loss function concerning the converging trend of a predicted sequence of each task in the corresponding iteration. We exploit a gating mechanism based on Gumbel distribution to learn an optimal way of incorporating the feedback signals. By applying our method to several baseline models in spoken language understanding, we demonstrate its effectiveness in improving model performance. Our work highlights the importance of incorporating posterior information into MTL and provides a promising direction for future research.


\bibliographystyle{IEEEbib}
\bibliography{refs}

\end{document}